\documentclass[10pt,twocolumn,letterpaper]{article}

\usepackage[pagenumbers]{cvpr} % To force page numbers, e.g. for an arXiv version

\usepackage[dvipsnames]{xcolor}

\definecolor{citeblue}{rgb}{0.21,0.49,0.74}
\usepackage[pagebackref,breaklinks,colorlinks,citecolor=citeblue]{hyperref}
\usepackage{multirow}
\usepackage{makecell}
\usepackage{tabularx}
\usepackage[symbol]{footmisc}
\usepackage{pifont}
\newcommand{\xmark}{\ding{55}}%
\newcommand{\cmark}{\ding{51}}%

\title{Insect-Foundation: A Foundation Model and Large-scale 1M Dataset for \\ Visual Insect Understanding}

\author{Hoang-Quan Nguyen$^{1}$\footnotemark[1]
, Thanh-Dat Truong$^{1}$\footnotemark[1]
, Xuan Bac Nguyen$^1$, Ashley Dowling$^2$, Xin Li$^3$, Khoa Luu$^1$\\
$^1$Department of Electrical Engineering and Computer Science, University of Arkansas, AR\\
$^2$Department of Entomology and Plant Pathology, University of Arkansas, AR\\
$^3$Department of Computer Science, SUNY Albany, NY \\
\tt\small{\{hn016, tt032, xnguyen, adowling, khoaluu\}@uark.edu, xli48@albany.edu} \\
\tt\small\url{https://uark-cviu.github.io/projects/insect_foundation.html} 
}

\begin{document}
\maketitle

\footnotetext[1]{Co-first authors}

\begin{abstract}

In precision agriculture, the detection and recognition of insects play an essential role in the ability of crops to grow healthy and produce a high-quality yield.  
The current machine vision model requires a large volume of data to achieve high performance. 
However, there are approximately 5.5 million different insect species in the world. None of the existing insect datasets can cover even a fraction of them due to varying geographic locations and acquisition costs.
In this paper, we introduce a novel ``Insect-1M'' dataset, a game-changing resource poised to revolutionize insect-related foundation model training.
Covering a vast spectrum of insect species, our dataset, including 1 million images with dense identification labels of taxonomy hierarchy and insect descriptions, offers a panoramic view of entomology, enabling foundation models to comprehend visual and semantic information about insects like never before.
Then, to efficiently establish an Insect Foundation Model, we develop a micro-feature self-supervised learning method with a Patch-wise Relevant Attention mechanism capable of discerning the subtle differences among insect images. In addition, we introduce Description Consistency loss to improve micro-feature modeling via insect descriptions.
Through our experiments, we illustrate the effectiveness of our proposed approach in insect modeling and achieve State-of-the-Art performance on standard benchmarks of insect-related tasks.
Our Insect Foundation Model and Dataset promise to empower the next generation of insect-related vision models, bringing them closer to the ultimate goal of precision agriculture. 

\end{abstract}

\vspace{-6mm}
\section{Introduction}

\begin{figure}[!t]
\centering
\includegraphics[width=0.95\linewidth]{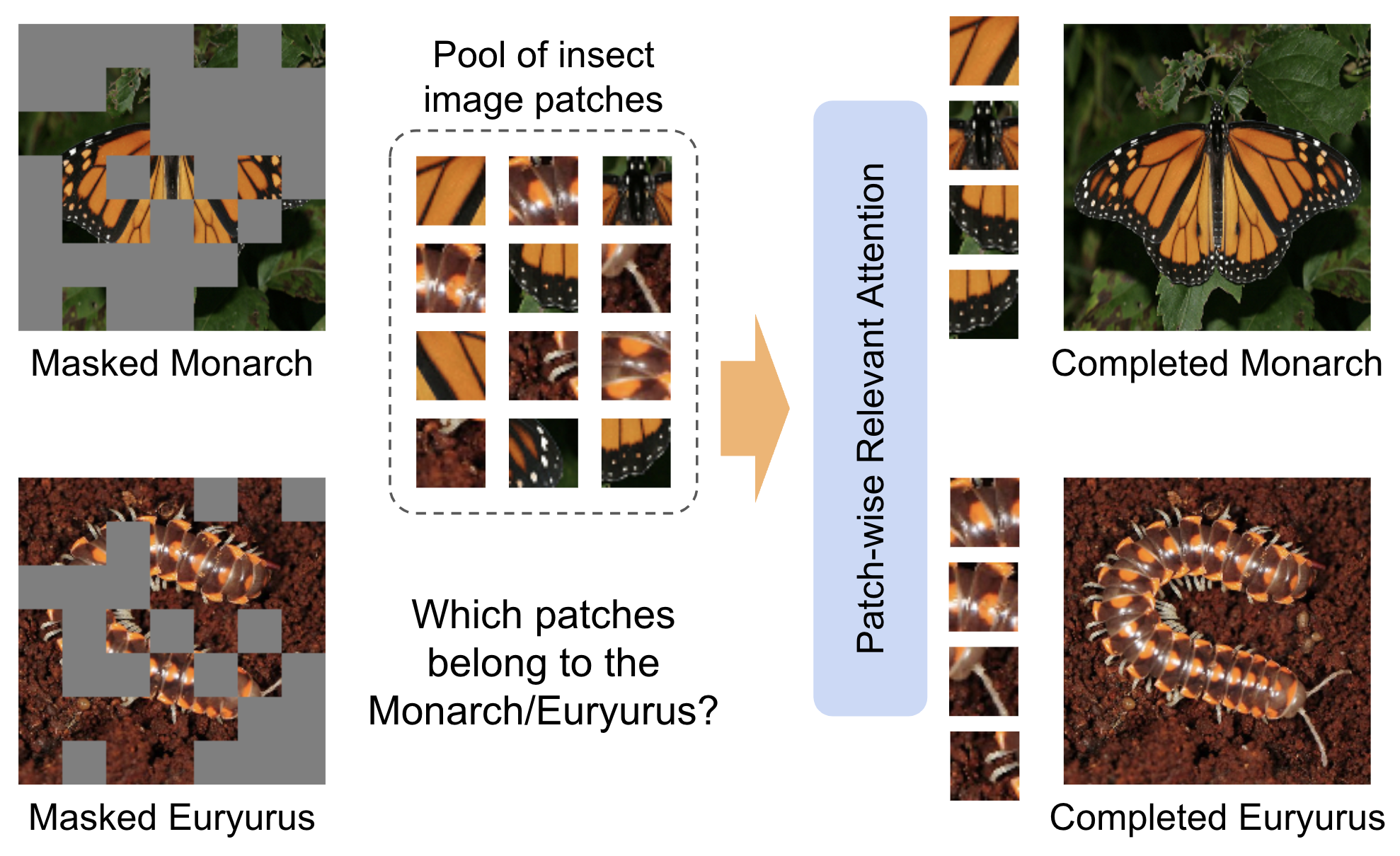}
\vspace{-4mm}
\caption{\textbf{Our Proposed Patch-wise Relevant Attention.}
Given masked insect images and separated image patches, our model can discriminate these patches that have small differences via relevant scores computed between masked images and image patches.}
\vspace{-7mm}
\label{fig:idea}
\end{figure}

Insects are the most diverse and abundant eukaryotic organisms on the planet. 
They inhabit all terrestrial and aquatic habitats and play a significant role within their community, habitat, and ecosystem as contributors to nutrient cycling, maintenance of plant and animal communities, disease cycling, and overall ecosystem health.
Therefore, in the agricultural revolution, the detection and identification of insects plays a key role in ensuring healthy crop growth and high-quality production.
Prior methods \cite{wu2019ip102,alves2020cotton,ayan2020crop,nanni2020insect,bollis2020weakly, truong2022otadapt} often fine-tuned the pre-trained ImageNet models on insect data for specific insect-related tasks, e.g., Insect Classification \cite{deng2018research,wu2019ip102,alves2020cotton,bollis2020weakly}, Insect Detection \cite{wu2019ip102}. However, these methods remained limited since the models pre-trained on ImageNet \cite{deng2009imagenet,krizhevsky2012imagenet,simonyan2014very,szegedy2015going,he2016deep,dosovitskiy2020image} could not model the micro features of insects, e.g., tiny texture and details of insects, as ImageNet \cite{deng2009imagenet} is the generic object dataset.

\begin{table*}[]
\centering
\caption{Comparison with existing datasets related to insects. Our proposed dataset has hierarchical labels with 6 main hierarchical levels, i.e., Subphylum, Class, Order, Family, Genus, and Species, and large numbers of species and samples. Moreover, the proposed dataset contains hierarchical descriptions for each insect and auxiliary taxonomic level, i.e., Subclass, Suborder, Subfamily, etc.}\label{tab:data_comparison}
\vspace{-3mm}
\resizebox{0.9\linewidth}{!}{
\begin{tabular}{lccccccc}
\Xhline{2\arrayrulewidth}
\textbf{Dataset} & \textbf{Year} & \textbf{Species} & \makecell{\textbf{Hierarchical}\\ \textbf{Labels}} & \makecell{\textbf{Hierarchical}\\ \textbf{Levels}} & \makecell{\textbf{Insect} \\ \textbf{Description}} & \makecell{\textbf{Auxiliary} \\ \textbf{Taxonomic Level}} & \makecell{\textbf{Number of} \\ \textbf{Samples}} \\
\hline
Samanta et al. \cite{samanta2012tea}       & 2012 & 8      & \xmark  & 1 & \xmark  & \xmark  & 609 \\
Wang et al. \cite{wang2012new}             & 2012 & 221      & \cmark  & 3 & \xmark  & \xmark & 225 \\
Venugoban et al. \cite{venugoban2014image} & 2014 & 20     & \xmark  & 1 & \xmark  & \xmark  & 200 \\
Xie et al. \cite{xie2015automatic}         & 2015 & 24     & \xmark  & 1 & \xmark  & \xmark & 1,440 \\
Liu et al. \cite{liu2016localization}      & 2016 & 12     & \xmark  & 1 & \xmark  & \xmark  & 5,136 \\
Xie et al. \cite{xie2018multi}             & 2018 & 40     & \xmark  & 1 & \xmark  & \xmark & 4,500 \\
Deng et al. \cite{deng2018research}        & 2018 & 10     & \xmark  & 1 & \xmark  & \xmark & 563 \\
Alfarisy et al. \cite{alfarisy2018deep}    & 2018 & 13     & \xmark  & 1 & \xmark  & \xmark  & 4,511 \\
PestNet \cite{liu2019pestnet} & 2019 & 16  & \xmark & 1 & \xmark & \xmark & 88,670 \\
IP102 \cite{wu2019ip102}                   & 2019 & 102    & \cmark & 3 & \xmark  & \xmark & 75,222 \\
AgriPest \cite{wang2021agripest} & 2021 & 14 & \cmark & 2 & \xmark & \xmark & 49,707 \\
INSECT \cite{badirli2021fine} & 2021 & 1,213 & \xmark & 1 & \xmark & \xmark & 21,212 \\
iNat-2021 \cite{van2021benchmarking} & 2021 & 2,752 & \cmark & 5 & \xmark & \xmark & 723,816 \\
\hline
\textbf{Our Insect-1M}                                       & \textbf{2023} & \textbf{34,212} & \textbf{\cmark} & \textbf{6} & \textbf{\cmark} & \textbf{\cmark} & \textbf{1,017,036} \\
\Xhline{2\arrayrulewidth}
\end{tabular}
}
\vspace{-6mm}
\end{table*}

Recent foundation models \cite{he2020momentum,chen2020improved,chen2020simple,xie2022simmim,he2022masked,caron2021emerging,oquab2023dinov2,radford2021learning,jia2021scaling,yu2022coca} pre-trained on large-scale datasets have revolutionized vision models with solid performance on downstream applications.
These models are designed to model general or specific properties of images or videos that can later be generalized to downstream tasks and unseen data.
The capability of the foundation model is often implemented with self-supervised or prompt-engineering training on large-scale datasets \cite{deng2009imagenet,jia2021scaling,zhai2022scaling,schuhmann2022laion}.
However, the current insect datasets \cite{deng2018research,wu2019ip102,alves2020cotton,bollis2020weakly,samanta2012tea,wang2012new,venugoban2014image,xie2015automatic,liu2016localization,xie2018multi,alfarisy2018deep} are insufficient to establish the foundation model of insects due to their scale and diversity.
Indeed, the most recent work presents an insect recognition dataset containing over $75,000$ images of 102 species \cite{wu2019ip102}.
Although the dataset includes many species, compared to the species of insects in the natural environment with over 5.5 million species \cite{stork2018many, ratnasingham2007bold}, the current work needs to have the diversity of insects.
Furthermore, to our knowledge, the current insect dataset \cite{wu2019ip102} does not provide the corresponding insect descriptions, limiting the ability to learn the foundation models.

Although the dataset is an important factor in developing an insect foundation model, the learning approach of the foundation model plays a significant role in performance.
There is significant progress in developing vision foundation models.
Common approaches learned alignment between vision and language, for example, CLIP \cite{radford2021learning}, ALIGN \cite{jia2021scaling}, CoCa \cite{yu2022coca}, to model visual concepts and data distributions.
Meanwhile, self-supervised contrastive or distillation learning approaches, e.g., MoCo \cite{he2020momentum,chen2020improved,chen2021empirical}, DINO \cite{caron2021emerging,oquab2023dinov2}, MAE \cite{he2022masked}, etc., learned the vision model by various pre-text tasks and have shown its scaling ability and generalizes well to various downstream tasks.
However, most of these previous foundation models represent the general information of natural images without specific knowledge.
When deploying in the insect domains, they cannot capture the micro-features of insects, i.e., key features or appearance to distinguish the species, since the texture and details of insects are often small and diverse compared to generic objects.
Meanwhile, fine-grained discrimination between insect images is crucial in insect foundation models due to the high diversity of species. 
Therefore, to successfully develop the insect foundation model, the learning approach needs to understand and be able to model the micro-features of insects.
Based on this observation, we present a novel pre-text task to enhance the recognition ability of the model between small features of the insect, as illustrated in Fig. \ref{fig:idea}.

\noindent
\textbf{Contributions of this Work:} To contribute to the development of the Insect Foundation Model in precision agriculture, we introduce a novel large-scale insect dataset, i.e., \textbf{\textit{Insect-1M}}, and a new Insect Foundation Model, i.e., \textbf{\textit{Insect-Foundation}}, that can transfer to various downstream insect-related applications, e.g., insect detection, insect classification, insect vision-language understanding.
Our contributions can be summarized as follows.
First, we present a new rich and large-volume insect dataset, i.e., \textbf{\textit{Insect-1M}}, that consists of 1 million images of insects with dense identifications of taxonomy hierarchy from the abstract level of taxonomy, e.g., Class, Order, to the detailed level of taxonomy, e.g., Genus, Species. In addition, each insect contains a detailed description that describes the details and features of insects. 
To the best of our knowledge, our proposed Insect-1M dataset is $13\times$ larger than the prior published IP102 dataset \cite{wu2019ip102}.
Second, to model the micro features of insects, we introduce a new self-supervised contrastive learning paradigm with a novel Patch-wise Relevant Attention mechanism to model the feature correlations of insect details.
Third, to increase the modeling capability of the Insect Foundation Model in learning insect details, we introduce a new Description Consistency loss to learn the detailed features of insects via the textual description.
Finally, through our intensive experiments on the Insect Classification and Insect Detection benchmarks \cite{wu2019ip102}, we show the effectiveness of our approach in insect modeling and our superior performance compared to the prior methods. 

\section{Related Work} 

\begin{figure*}[!t]
\centering
\includegraphics[width=0.8\linewidth]{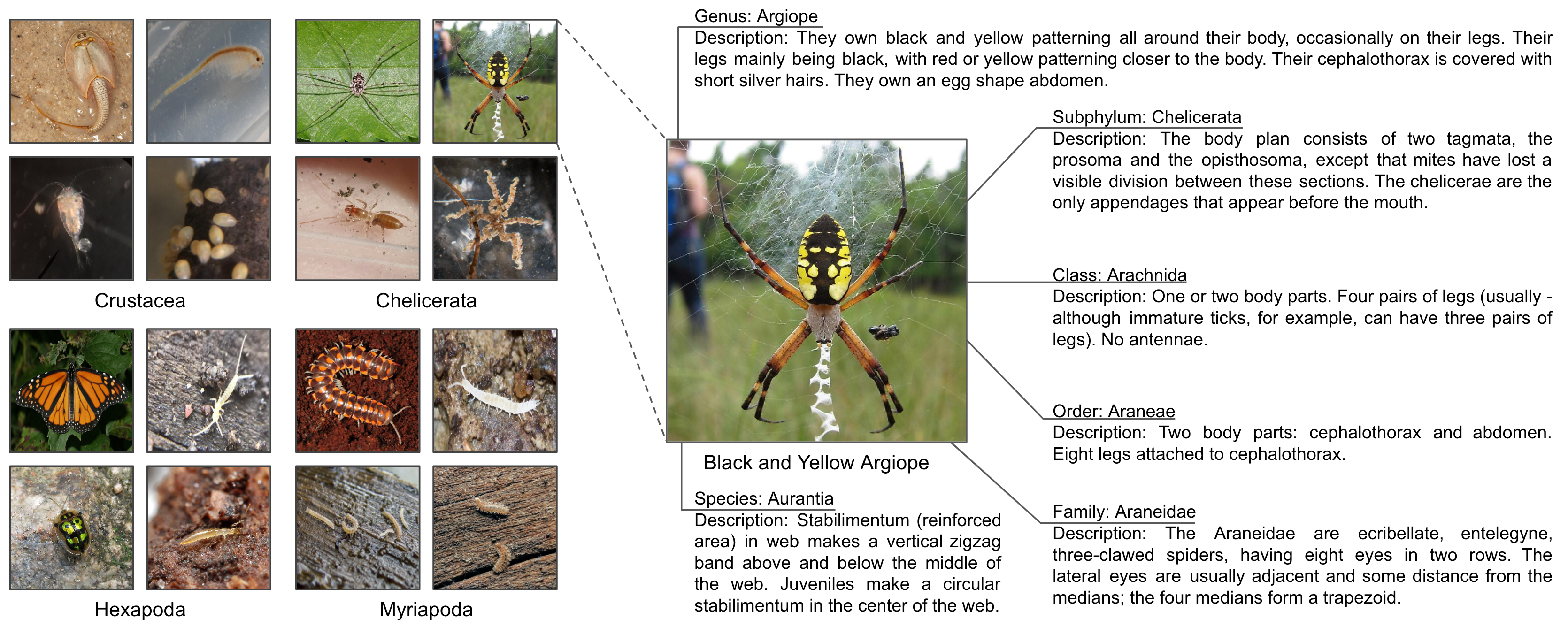}
\vspace{-3mm}
\caption{\textbf{Examples of Our Insect-1M Dataset.} The left figure illustrates the samples of the four Subphylums, including Chelicerata, Crustacea, Hexapoda, and Myriapoda. The right figure shows an example of hierarchical descriptions of the Aurantia Species.} 
\label{fig:subphylums_with_description}
\vspace{-6mm}
\end{figure*}

\noindent
\textbf{Insect Datasets.} There are prior studies releasing insect datasets on a small scale for recognition problems.
\cite{venugoban2014image} presented a dataset consisting of $20$ species with $10$ samples for each species. 
Then, \cite{xie2015automatic} introduced an insect dataset including $1,440$ samples of $24$ species.
Several subsequent studies have larger datasets for deep learning, e.g., \cite{xie2018multi} proposed an insect dataset of $4,500$ images with 40 different species for insect classification,
and \cite{liu2016localization} proposed an insect dataset with over $5,000$ samples for insect recognition and localization.
PestNet \cite{liu2019pestnet} and AgriPest \cite{wang2021agripest} were introduced for the small pest detection task.
Recently, \cite{wu2019ip102} has presented IP102 as a large-scale dataset containing over $75K$ samples of insects with $102$ species for classification and detection tasks.
Meanwhile, \cite{van2021benchmarking} proposed a large-scale dataset including over $723K$ samples of Arthropoda phylum with $2,752$ species.
Although prior efforts promoted the development of vision and machine intelligence in precision agriculture, no dataset has a large volume of samples and diverse species for insect-related foundation model training. 
Therefore, this work introduces a novel dataset that not only contains a large number of samples, i.e. 1M images, but also has hierarchical labels from the high to the low taxonomy level, including class, order, family, genus, and species.
Table \ref{tab:data_comparison}  compares our proposed dataset with the prior ones.
In comparison with prior datasets, the number of images in our proposed Insect-1M dataset is $13\times$ higher than the prior IP102 dataset, and the number of species is $335\times$ higher than IP102 \cite{wu2019ip102}.
To preserve the rights of datasets and authors of images, instead of publishing images, we only provide labels and links to download images.

\noindent
\textbf{Self-supervised Pre-training.}
Self-supervised pre-training has become a popular strategy for solving visual recognition problems, including classification, localization, segmentation, video recognition, tracking, and many other problems \cite{he2022masked, truong2022direcformer, nguyen2021clusformer, truong2021bimal, truong2024fairness, truong2023fredom, truong2023liaad,nguyen2023brainformer,nguyen2023fairness,sefl_supervised_medical,nguyen2023micron,nguyen2022two}.
SimCLR \cite{chen2020simple} learned the visual representation of images via a contrastive learning framework using different data augmentation operations.
MoCo \cite{he2020momentum} introduced momentum updating for the encoder while learning the image representation via contrastive learning.
The MoCo framework was later used to improve the SimCLR approach without requiring a large training batch size \cite{chen2020improved}.
MoCo-V3 \cite{chen2021empirical} improved prior Momentum Contrastive frameworks by eliminating the memory queue to stabilize the training when the batch size is large.
DINO \cite{caron2021emerging} proposed a self-supervised learning approach using knowledge distillation with no labels. Later, it was extended to DINO-V2 \cite{oquab2023dinov2} by stabilizing self-supervised learning when scaling the size of models and data.
BEiT \cite{bao2021beit} proposed a masked image modeling task and used discrete visual tokens from the original image as prediction targets.
MAE \cite{he2022masked} and SimMIM \cite{xie2022simmim} directly used a decoder to reconstruct pixel values from masked regions.
Jigsaw-ViT \cite{chen2023jigsaw} presented a pre-training task for transformer models by solving the shuffled patches of images. 
This learning strategy was also applied on the temporal dimension to improve the robustness of video modeling \cite{truong2022direcformer}.
Micron-BERT \cite{nguyen2023micron} studied the micro-changing in facial videos by learning to detect the minor differences in an image that has swapping regions between two frames.

\noindent
\textbf{Joint Vision-Language Pre-training.}
Recent work introduced joint vision-language pre-training.
CLIP \cite{radford2021learning}, and ALIGN \cite{jia2021scaling} addressed that dual-encoder models pre-trained on image-text pairs in contrastive objectives can learn strong representations of image and text for cross-modal alignment and zero-shot image recognition problems.
LiT \cite{zhai2022lit} and BASIC \cite{pham2023combined} proposed zero-shot transfer learning approaches by teaching the text model to learn the representation of the pre-trained image model via contrastive losses with large-scale data.
SimVLM \cite{wang2021simvlm}, OFA \cite{wang2022ofa}, and BLIP \cite{li2022blip} trained an encoder-decoder model with language generative losses and achieved high performance in the vision-language benchmarks.
CoCa \cite{yu2022coca} utilized contrastive learning and generative image captioning for global representation learning and fine-grained image-text alignment.
Later work \cite{zhai2023sigmoid} used sigmoid loss to compute the image-text similarity for batch size scaling.
LexLIP \cite{luo2023lexlip} projected images into a lexicon space for image-text sparse matching.
Meanwhile, EQSIM \cite{wang2023equivariant} computed the similarity by the image-text equivariant changing.

\begin{figure*}[!t]
\begin{center}
\includegraphics[width=0.8\linewidth]{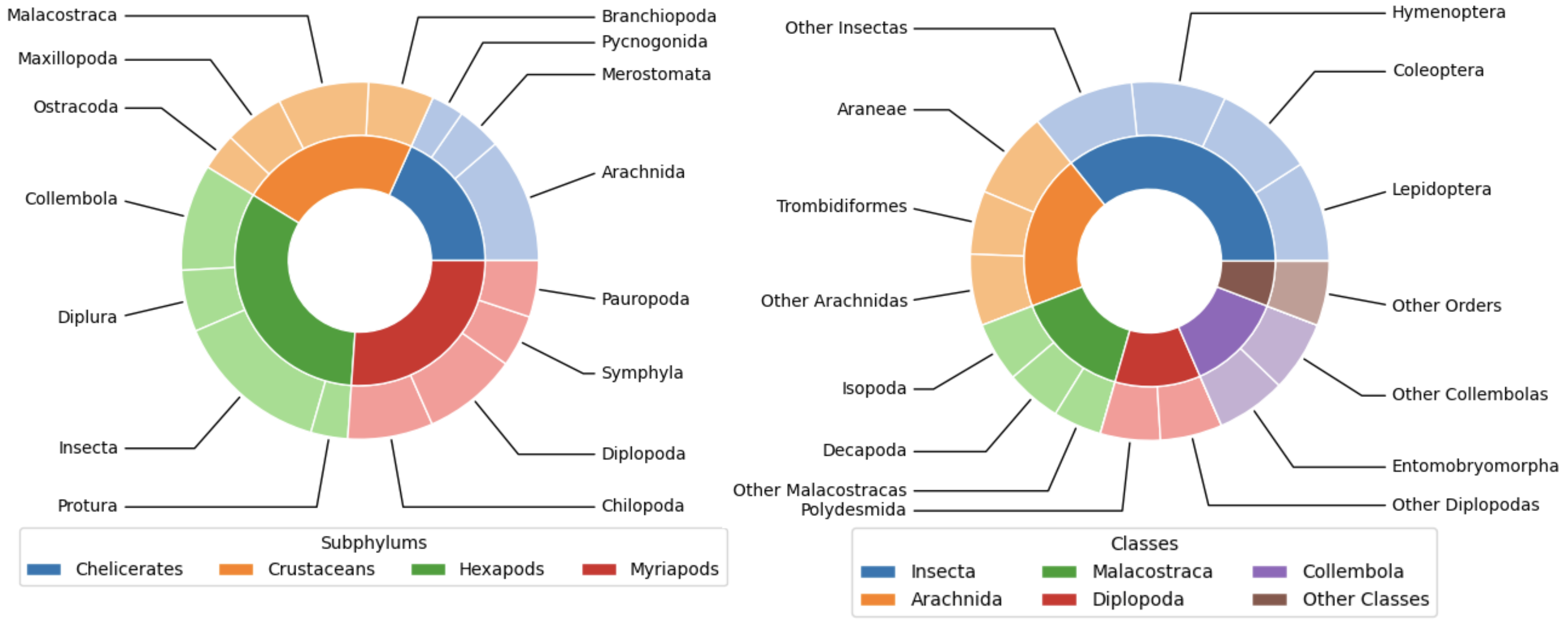}
\end{center}
\vspace{-7mm}
\caption{The Distribution of Subphylum and Its Classes (Left) and The Distribution of Class and Its Orders (Right). \textbf{Best viewed in color.}}
\label{fig:data_chart}
\vspace{-6mm}
\end{figure*}

\section{The Proposed Insect 1M Dataset}

To contribute to establishing the insect foundation model, the large-scale dataset of insects with diverse species is essential. Therefore, we collect a new insect dataset with dense labels of a hierarchical taxonomy. In particular, our Insect-1M dataset contains 1 million insect images with dense hierarchical labels with six main taxonomies, i.e., Subphylum, Class\footnote{In this paper, we use the term ``Class'' as a biological taxonomic level.}, Order, Family, Genus, and Species.
The samples are in the Phylum Arthropoda and can be divided into 4 Subphylums, which are Chelicerata, Crustacea, Hexapoda, and Myriapoda as shown in Fig. \ref{fig:subphylums_with_description}. 
Compared to prior datasets, our Insect-1M has more hierarchical levels with large numbers of species and samples as in Table \ref{tab:data_comparison}.

\subsection{Data Collection Protocol}

We utilize insect information containing insect data with images and taxonomies collected by naturalists and entomologists.
Each insect sample has a corresponding image and its taxonomic label.
From the taxonomic label, we crawl the identification description of the corresponding taxonomy.
Notice that the taxonomic labels are hierarchical. The description is written from high-level descriptions, e.g., Subphylum and Class, to low-level descriptions, e.g., Species. Fig. \ref{fig:subphylums_with_description} shows an example of an insect description.

\subsection{Data Preprocessing and Statistic}

\noindent
\textbf{Data Preprocessing.} The raw data is stored in over 1 million HTML files with predefined HTML structures.
Then, we parse the data structures to collect the insect images and their labels.
More than 2 million raw images and their corresponding labels have been collected. 
However, the raw data collected consists of a lot of noise, e.g., incorrect identification of insects, corrupted images, and non-insect images. Therefore, to filter these outliers, our entomology experts must verify the images and their labels, i.e., insect identification. 
Finally, our collected Insect-1M dataset consists of $1,017,036$ clean images with dense labels of $34,212$ different insect species.

\noindent
\textbf{Data Statistic}
Fig. \ref{fig:data_chart} 
shows the sample distributions of the Subphylums and their Classes.
It is shown that the Class Insecta has the majority of samples.
Fig. \ref{fig:data_chart} also illustrates the distribution of the Orders in the major Classes.
For each major Class, the data distribution of Orders is well-balanced.

\section{The Proposed Insect Foundation Model}

\begin{figure}[!b]
\vspace{-6mm}
\centering
\includegraphics[width=0.95\linewidth]{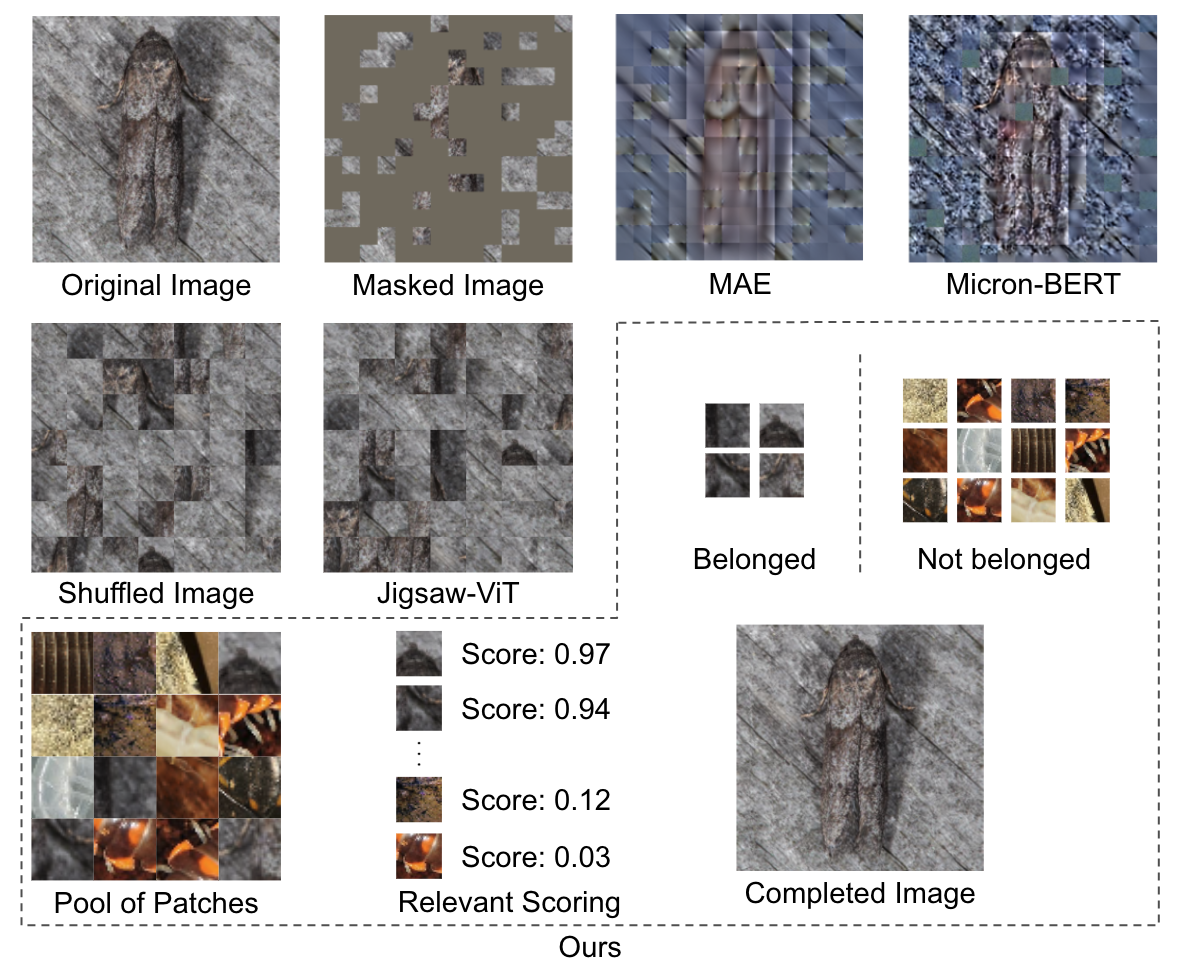}
\vspace{-4mm}
\caption{\textbf{Comparisons of Self-supervised Methods.} MAE \cite{he2022masked} fails to reconstruct the details of the insect since it learns general information about the image. Micron-BERT \cite{nguyen2023micron} hardly distinguishes the insect and background. Jigsaw-ViT \cite{chen2023jigsaw} cannot correct shuffled patches due to confusion between the background and the object. Meanwhile, our approach can find separated patches belonging to the insect by scoring each patch.
\textbf{Best viewed in color.}
}
\label{fig:methods_comparison}
\end{figure}
\subsection{Limitations of Prior Foundation Training Approaches}

\begin{figure*}[!t]
\begin{center}
\includegraphics[width=0.9\linewidth]{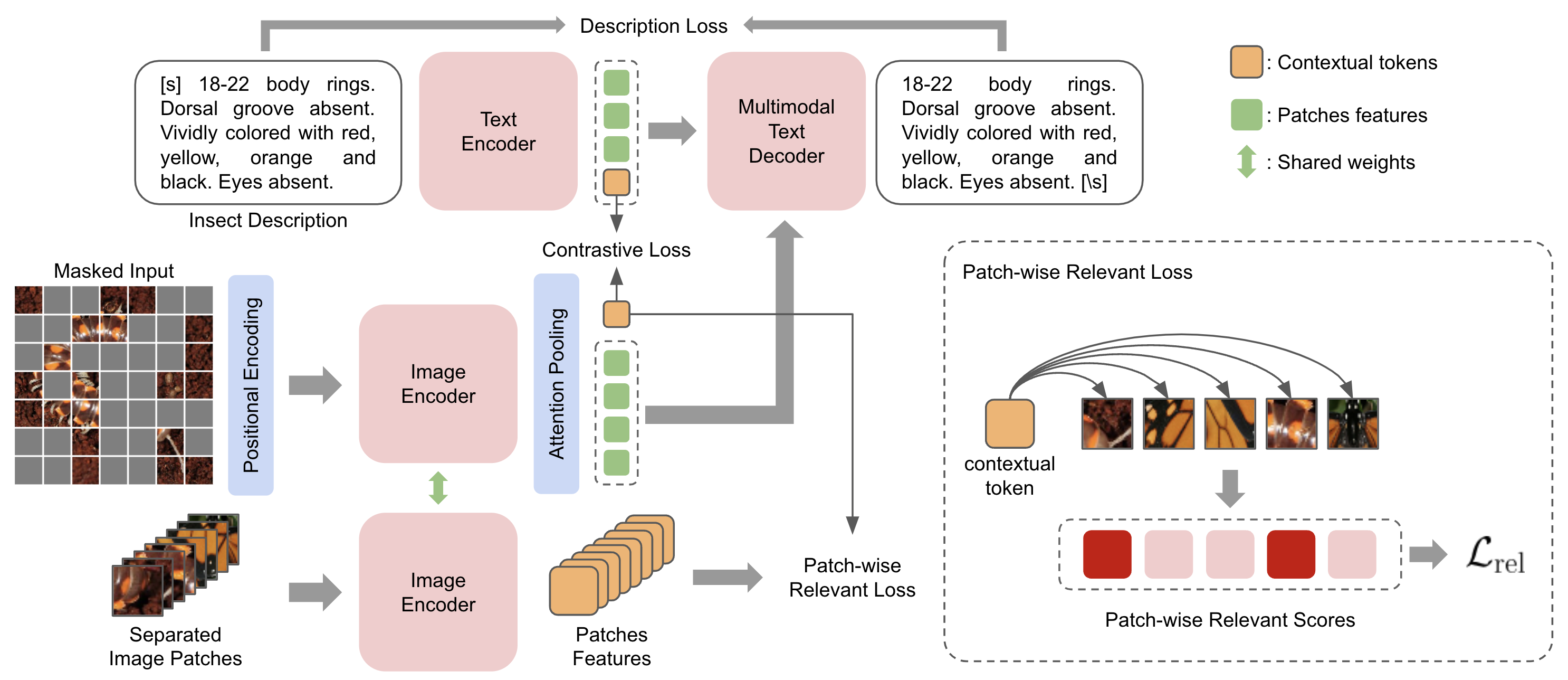}
\end{center}
\vspace{-6mm}
\caption{The Overview Framework of Our Proposed Approach to Insect Foundation Model.}
\label{fig:overview_framework}
\vspace{-6mm}
\end{figure*}

\textbf{Limitations} One of the issues in the visual insect understanding problem is the visual representation and discrimination of the small and undistinguished features of the insects.
While MAE \cite{he2022masked} reconstructs an image from a masked image for visual representation learning, it focuses on the context inside the image individually without realizing the small details to discriminate between the insects. 
Meanwhile, Jigsaw solving methods \cite{noroozi2016unsupervised,chen2023jigsaw} correct the position of image patches to enhance the model robustness to the image structure. This strategy needs more mechanisms to focus on the small details of the image.
Micron-BERT \cite{nguyen2023micron} highlights the small changes in the image by swapping the regions between two images with similar contexts.
However, the small changes in the insect image still preserve the signature features representing the insect.
Thus, it makes the model collapse in detecting the small features of insects.
Therefore, to address these limitations, we introduce a new approach that learns to recognize the tiny features in the insect images.
These features are distinguished from the background by discriminating the minor differences between patches of images individually.
Fig. \ref{fig:methods_comparison} compares prior self-supervised methods \cite{he2022masked,nguyen2023micron,chen2023jigsaw} with our approach.

Fig. \ref{fig:overview_framework} illustrates our insect foundation model.
The model is designed to capture the small differences in insect features, i.e., textures or limbs, via our new self-supervised pre-text task.
Moreover, the model is pre-trained to learn the fine-grained alignment between the insect description and its visual features.
Formally, given an input image $I$, we divide $I$ into non-overlapping patches.
Then, a subset of patches $P_s$ is sampled, and the remaining patches are put into a pool of image patches $P_{\text{pool}}$.
The sampling is processed randomly in a uniform distribution.
An image encoder is used to map $I_p$ into latent vectors.
Given an insect description $T$ of the image, a text encoder is presented to extract information from $T$.
A text decoder and joint image-text contrastive learning module are introduced to map the description into the image.
Finally, a Patch-wise Relevant Attention module is proposed for self-supervised learning to enhance the discrimination robustness of the model.

\subsection{Input Modeling}

An input image $I \in \mathbb{R}^{H \times W \times 3}$  is divided into non-overlapping patches $P = \{p_s^i\}_{i=1}^{N_P}$ where $H, W$ are the height and width of the input image, $N_P = HW/(s_p)^2$ is the number of patches. 
Each patch $p_s^i$ has a resolution of $s_p \times s_p$.
The non-overlapping patches $P$ are then randomly sampled into a subset of patches $P_s \subset P$ and put the other patches into a pool of image patches $P_{\text{pool}}$. 
Note that $P_{\text{pool}}$ contains patches from multiple images in the training set.

\subsection{Image Encoder}

Each patch $p_s^i \in P_s$ is projected into a latent vector $\mathbf{x}_s^i \in \mathbb{R}^d$ where $d$ is the dimension of the latent vectors.
A subset patches $P_s$ can be represented as follows:
\vspace{-2mm}
\begin{equation}
\small
    \mathbf{X}_s = \text{concat}[\mathbf{x}_s^i]_{i=1}^{N_{P_s}} \in \mathbb{R}^{N_{P_s} \times d}, \quad
    \mathbf{x}_s^i = \alpha_p(p_s^i) + \mathbf{e}_p(i)
\vspace{-2mm}
\end{equation}
where $\alpha_p$ and $\mathbf{e}_p$ are the projection embedding and position embedding.

Let an image encoder $E_{\text{image}}(\mathbf{X}_s)$ be a stack of $L_e$ transformer blocks where each block contains multi-head self-attention (MSA) and multi-layer perceptron (MLP).
\vspace{-2mm}
\begin{equation}
\vspace{-2mm}
\small
\begin{split}
    \mathbf{X}^\prime_l &= \mathbf{X}_{l-1} + \text{MSA}(\text{LN}(\mathbf{X}_{l-1})) \\
    \mathbf{X}_l &= \mathbf{X}^\prime_l + \text{MLP}(\text{LN}(\mathbf{X}^\prime_l)) \\
    \mathbf{X}_0 &= \mathbf{X}_s, \: 1 \leq l \leq L_e
\end{split}
\end{equation}
where $\text{LN}$ is the layer normalization. 
Then, given $\mathbf{X}_s$, the output latent vector $\mathbf{Z}_s$ is represented as follows:
\vspace{-2mm}
\begin{equation}
\small
    \mathbf{Z}_s = E_{\text{image}}(\mathbf{X}_s), \quad \mathbf{Z}_s \in \mathbb{R}^{N_{P_s} \times d}
\end{equation}

\subsection{Insect Micro-feature Self-supervised Learning}

\begin{figure}[!b]
\begin{center}
\includegraphics[width=0.85\linewidth]{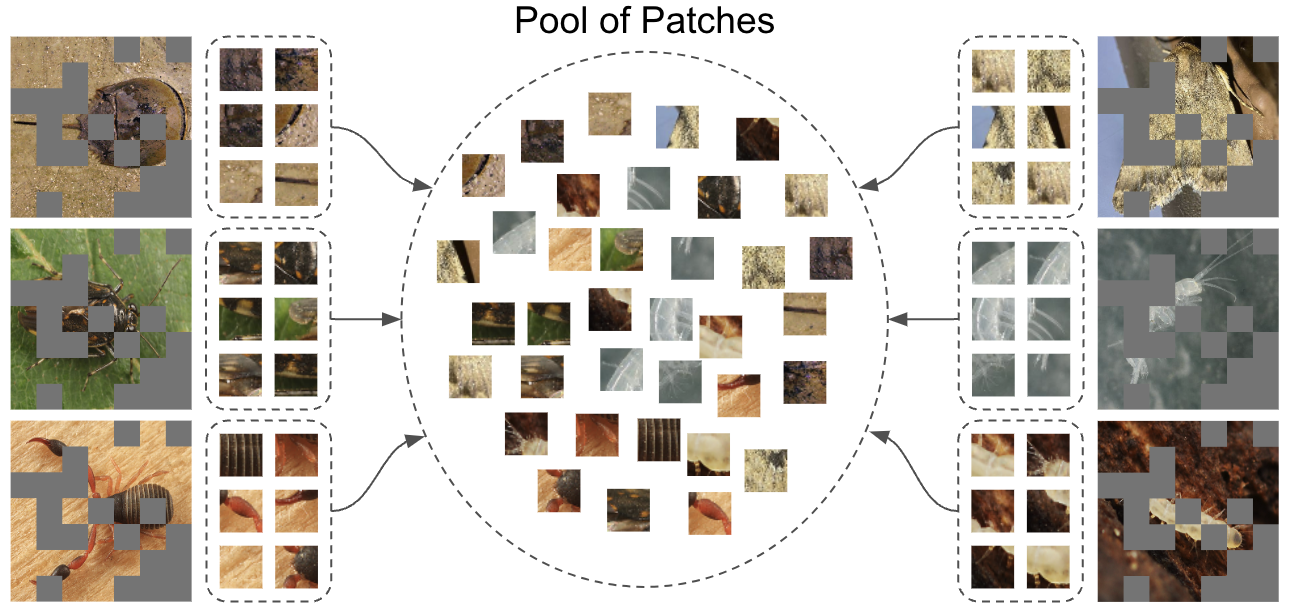}
\end{center}
\vspace{-6mm}
\caption{\textbf{Pool of Image Patches.} A subset of patches of an image is sampled for image encoding while the remaining patches are placed into a pool of patches for the self-supervised pre-text task.}
\label{fig:pool_of_patches}
\end{figure}

The recognition of insects relies on the insect texture, eyes, or limbs that are tiny to detect.
To make the model robust to the small features of insect images, we propose a self-supervised learning strategy to spot these small features via the small differences in the images.
Notice that the insects can be distinguished by detecting and discriminating the critical features in each part of those insects. 
To enhance this ability for the model, a pre-text task is presented. 
In particular, after extracting global information from a masked image of the insect, the vision model learns to find the remaining patches of the image by comparing image patches of different insect species. 
Thanks to our learning mechanism, the model learns the key features representing each insect and discriminates the small features between different species.
As illustrated in Fig. \ref{fig:pool_of_patches}, given a subset of patches $P_s$ from the image $I$ and a pool of image patches $P_{\text{pool}}$, we train the model to find the patches $p_t \in P_{\text{pool}}$ that originally belong to the image $I$.
Then, given latent vectors $\mathbf{Z}_s$ of $P_s$, a patch-wise relevant attention score (PRS) is computed between $\mathbf{Z}_s$ and each patch $p \in P_{\text{pool}}$.
The score can be defined as:
\vspace{-1mm}
\begin{equation}
\vspace{-1mm}
\small
    \text{PRS} = f(\mathbf{Z}_s, p) \in [0, 1]
\label{eq:prs_abstract}
\end{equation}
The higher the score is, the more possibility that $p \in P$.

\begin{figure}[!b]
\begin{center}
\includegraphics[width=0.85\linewidth]{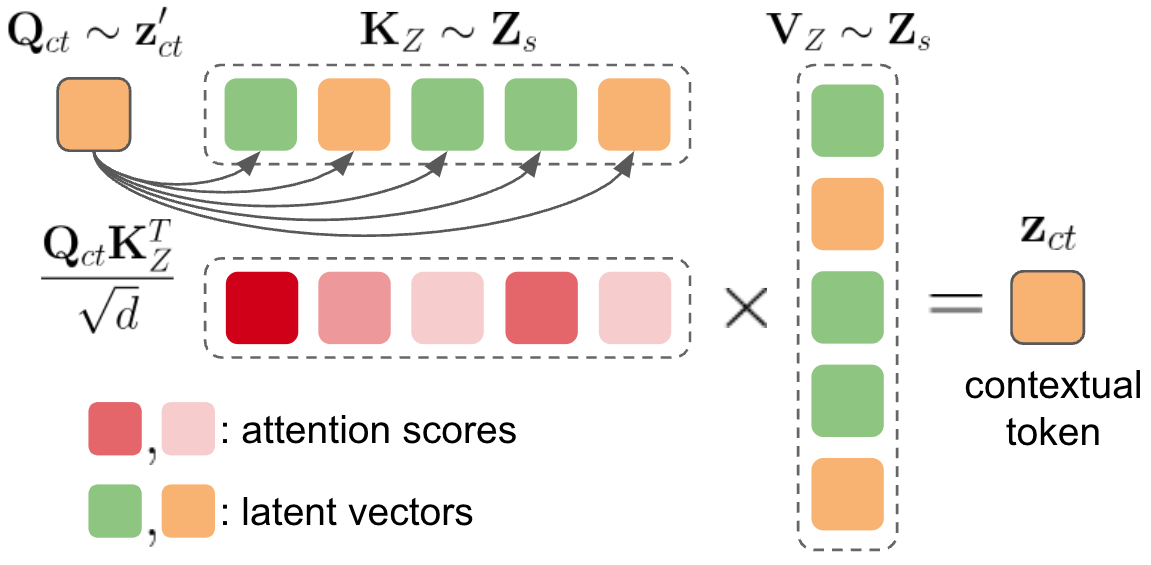}
\end{center}
\vspace{-6mm}
\caption{\textbf{Attention Pooling Module.} The contextual token $\mathbf{z}_{ct}$ represents the global information of the image $I$.}
\label{fig:attention_pooling}
\end{figure}

\noindent
\textbf{Attention Pooling}
To compute the relevance between latent vectors $\mathbf{Z}_s$ from the image $I$ and the patch $p \in P_{\text{pool}}$, the latent vectors $\mathbf{Z}_s$ should be aggregated to represent the holistic information of $I$.
Inspired by \cite{yu2022coca}, we compute the global information of $I$ via attention pooling.
Given a placeholder contextual token $\mathbf{z}_{ct}^\prime$ as a query $\mathbf{Q}_{ct}$ and latent vectors $\mathbf{Z}_s$ as a key $\mathbf{K}_{Z}$ and a value $\mathbf{V}_{Z}$, we compute an attention map between $\mathbf{Q}_{ct}$ and $\mathbf{K}_{Z}$. 
Then, a contextual token $\mathbf{z}_{ct}$ representing the global information of $I$ is computed via the attention map and the value $\mathbf{V}_{Z}$.
The attention pooling (Fig. \ref{fig:attention_pooling}) can be formulated as Eqn. \eqref{eq:attn_pool}.
\begin{equation}
\vspace{-1mm}
\small
\begin{split}
    \mathbf{Q}_{ct} &= \text{Linear}(\mathbf{z}_{ct}^\prime) \quad \mathbf{K}_{Z} = \text{Linear}(\mathbf{Z}_s) \quad \mathbf{V}_{Z} = \text{Linear}(\mathbf{Z}_s) \\
    \mathbf{z}_{ct} &= \text{softmax}\left(\frac{\mathbf{Q}_{ct} \mathbf{K}_{Z}^T}{\sqrt{d}}\right) \mathbf{V}_{Z}
\end{split}
\label{eq:attn_pool}
\end{equation}

\noindent
\textbf{Patch-wise Relevant Attention}
Given $\mathbf{z}_{ct}$ as a contextual token representing the information of $I$, we compute the relevance between $\mathbf{z}_{ct}$ and $p \in P_\text{pool}$.
From Eqn. \eqref{eq:prs_abstract}, we expand the attention score function $f$ as in Eqn. \eqref{eq:prs}.
\vspace{-1mm}
\begin{equation}
\vspace{-1mm}
\small
    \text{PRS} = f(\mathbf{Z}_s, p) = H(\mathbf{z}_{ct}, \mathbf{z}_p)
\label{eq:prs}
\end{equation}
where $\mathbf{z}_p = E_\text{image}(\alpha_p(p))$ is a latent vector representing the patch $p$, $H$ is a similarity function between two latent vectors.
From Eqn. \eqref{eq:prs}, we expand the score function into a self-supervised loss function $\mathcal{L}_{\text{PRS}}$ as follow:
\vspace{-1mm}
\begin{equation}
\vspace{-1mm}
\small
    \mathcal{L}_{\text{rel}} = - y \log(H(\mathbf{z}_{ct}, \mathbf{z}_p)) - (1 - y) \log(1 - H(\mathbf{z}_{ct}, \mathbf{z}_p))
\end{equation}
where $y = 1$ if $p \in P$ and $y = 0$ otherwise.

\subsection{Fine-grained Insect Image-Text Alignment}

Each species has an individual definition and description that can be aligned to parts of the insect image.
We adopt a text decoder to generate the species descriptions from insect images.
Moreover, to capture the general information of species, we utilize contrastive learning between global features of the insect images and description.
As a result, the model can learn specific information from insect images via insect descriptions.

Formally, an insect description text is tokenized into $T = \{t_i\}_{i=1}^{N_T}$ where $N_T$ is the number of tokens of the description.
Each token $t_i \in T$ is embedded into a latent vector $\mathbf{w}_i \in \mathbb{R}^d$.
The description can be represented as:
\vspace{-1mm}
\begin{equation}
\vspace{-1mm}
\small
    \mathbf{W} = \text{concat}[\mathbf{w}_i]_{i=1}^{N_T} \in \mathbb{R}^{N_T \times d}, \quad
    \mathbf{w}_i = \alpha_w + \mathbf{e}_w(i)
\label{eq:text_input_model}
\end{equation}
where $\alpha_w$ and $\mathbf{e}_w$ are the projection embedding and position embedding.

Similar to the image encoder, let the text encoder $E_\text{text}(\mathbf{W})$ be a stack of $L^\prime_e$ transformer blocks containing multi-head self-attention and multi-layer perceptron.
The output latent vector $\mathbf{Z}^\prime$ of the description is computed as
\vspace{-1mm}
\begin{equation}
\vspace{-1mm}
\small
    \mathbf{W}^\prime = E_\text{text}(\mathbf{W}), \quad \mathbf{Z}^\prime \in \mathbb{R}^{N_T \times d}
\end{equation}
We then use the latent vector $\mathbf{Z}_s$ of the insect image and $\mathbf{W}^\prime$ of the description text for image-text contrastive learning and multi-modal image description decoding.

\noindent
\textbf{Image-text Contrastive Learning.}
Inspired by the prior language model frameworks \cite{devlin2018bert,liu2019roberta,lewis2019bart,raffel2020exploring}, a contextual token $\mathbf{w}_{ct}$ representing the semantic information of the description is added at the beginning of $\mathbf{W}$ as in Eqn. \ref{eq:text_input_model}.
Then the two encoders $E_\text{image}$ and $E_\text{text}$ can be jointly optimized via contrastive learning as follow:
\vspace{-1mm}
\begin{equation}
\vspace{-1mm}
\footnotesize
\begin{split}
    \mathcal{L}_\text{con} &= \frac{-1}{N} \sum_{i=1}^N \left[\log 
    \frac{\exp( \mathbf{z}_i^T \mathbf{w}_i )}{\sum_{j=1}^N \exp( \mathbf{z}_i^T \mathbf{w}_j )} + \log 
    \frac{\exp( \mathbf{w}_i^T \mathbf{z}_i )}{\sum_{j=1}^N \exp( \mathbf{w}_i^T \mathbf{z}_j )}\right]
\end{split}
\label{eq:contrastive_text_image}
\end{equation}
where $\mathbf{z}_i$ and $\mathbf{w}_i$ is the contextual token of the $i$-th insect image and description.

\noindent
\textbf{Multi-modal Image Description Decoding.}
While image-text contrastive learning represents the global semantic information between the image and description, the multi-model image description decoding aims for the fine-grained details by predicting the tokenized texts of $T$ in an autoregressive manner, as shown in Eqn. \eqref{eq:description_loss}.
\begin{equation}
\small
    \mathcal{L}_\text{desc} = 
    - \sum_{t=1}^{N_T} \log D_\text{multi} (\mathbf{w}_t | \mathbf{W}_{0:t-1}, \mathbf{Z}_s)
\label{eq:description_loss}
\vspace{-1mm}
\end{equation}
where $D_\text{multi}$ is an autoregressive multi-modal text decoder.

\section{Experimental Results}

\subsection{Foundation Model Pre-training}

Our experiments use ViT-Base (ViT-B/16) \cite{dosovitskiy2020image} as the backbone.
The images are resized and cropped randomly into the resolution of $224 \times 224$.
Then, each image is divided into patches of $16 \times 16$, creating $N_P = 196$ patches.
The patch sampling ratio is selected as $50\%$, and the remaining patches are put into the pool of image patches.
Each patch is projected to latent space of $d = 768$ dimensions.
The text encoder and multi-modal text decoder are adopted from the pre-trained BERT model \cite{devlin2018bert}.
The model is implemented in PyTorch \cite{paszke2019pytorch} and trained by $16 \times \text{A100}$ GPUs.
The learning rate is initially set to $1.5 \times 10^{-4}$ with the Consine learning rate scheduler \cite{loshchilov2016sgdr}.
The model is optimized by AdamW \cite{loshchilov2017decoupled} with 200 epochs and a batch size of 64 per GPU.

\begin{table}[t]
\centering
\caption{\textbf{Effectiveness of our method on the IP102 Classification.} We evaluate approach with three different vision transformer backbones, i.e., ViT-small/16, ViT-base/16, and ViT-large/16, without or with Attention Pooling (Attn Pool), and three different losses, i.e. Patch-wise Relevant Loss ($\mathcal{L}_\text{rel}$), Image-Text Contrastive Loss ($\mathcal{L}_\text{con}$), and Description Loss ($\mathcal{L}_\text{desc}$).}
\vspace{-3mm}
\resizebox{0.85\linewidth}{!}{
\begin{tabular}{c|cccc|cc}
\toprule
\textbf{Backbone} &
  $\mathcal{L}_\text{rel}$ &
  \makecell{\textbf{Attn} \\ \textbf{Pool}} &
  $\mathcal{L}_\text{con}$ &
  $\mathcal{L}_\text{desc}$ &
  \makecell{\textbf{Acc@1} \\ \textbf{(\%)}} &
  \makecell{\textbf{Acc@5} \\ \textbf{(\%)}} \\
% \hline
\midrule
\multirow{4}{*}{ViT-small/16} & \checkmark &            &            &            & 68.9 & 88.8 \\
                              & \checkmark & \checkmark &            &            & 69.5 & 89.7 \\
                              & \checkmark & \checkmark & \checkmark &            & 70.7 & \textbf{89.9} \\
                              & \checkmark & \checkmark & \checkmark & \checkmark & \textbf{71.5} & 87.7 \\
% \hline
\midrule
\multirow{4}{*}{ViT-base/16}  & \checkmark &            &            &            & 72.4 & 91.0 \\
                              & \checkmark & \checkmark &            &            & 73.3 & 91.6 \\
                              & \checkmark & \checkmark & \checkmark &            & 74.2 & 91.9 \\
                              & \checkmark & \checkmark & \checkmark & \checkmark & \textbf{75.8} & \textbf{92.1} \\
% \hline
\midrule
\multirow{4}{*}{ViT-large/16} & \checkmark &            &            &            & 73.8 & 90.9 \\
                              & \checkmark & \checkmark &            &            & 74.6 & 91.6 \\
                              & \checkmark & \checkmark & \checkmark &            & 75.9 & 91.4 \\
                              & \checkmark & \checkmark & \checkmark & \checkmark & \textbf{76.9} & \textbf{92.7} \\
% \Xhline{2\arrayrulewidth}
\bottomrule
\end{tabular}
}
\label{tab:abl_studies}
\vspace{-6mm}
\end{table}

\subsection{Datasets and Benchmarks}

\noindent
\textbf{IP102 Classification} \cite{wu2019ip102} provides 102 species of insects and contains 45,095 training samples, 7,508 validation samples, and 22,619 testing
samples.
For each species, an image might contain a single insect, multiple insects, or even a diseased crop caused by the species.
The insects are in different forms for each class, e.g., egg, larva, pupa, and adult. 
The performance of insect classification is evaluated by the accuracy of Top 1 (Acc@1) and Top 5 (Acc@5).

\noindent
\textbf{IP102 Detection} \cite{wu2019ip102} includes 15,178 training images and 3,798 testing images of 102 different species.
Following the COCO benchmark \cite{lin2014microsoft}, the insect detection performance is measured by the Average Precision (AP) and Average Precision at IoU thresholds of 0.5 (AP$^{.50}$) and 0.75 (AP$^{.75}$).

\subsection{Ablation Studies}

Our ablation experiments study the effectiveness of our proposed model and hyper-parameters on the IP102 Classification Benchmark as shown in Table \ref{tab:abl_studies}.

\begin{table}[!t]
\centering
\caption{\textbf{Classification results on IP102 Classification benchmark.} Both proposed models pre-trained with and without the insect descriptions outperform prior methods by a large margin.}\label{tab:ip102_classification}
\vspace{-3mm}
\resizebox{0.4\textwidth}{!}{
\begin{tabular}{lcccc}
\Xhline{2\arrayrulewidth}
\textbf{Method}  & \textbf{Description} & \makecell{\textbf{Pre-train} \\  \textbf{Data}}  & \makecell{\textbf{Acc@1} \\ \textbf{(\%)}} & \makecell{\textbf{Acc@5} \\ \textbf{(\%)}}  \\
\hline
ResNet \cite{wu2019ip102} & \xmark & ImageNet1K & 49.4 & - \\
EfficientNet \cite{bollis2020weakly} & \xmark & ImageNet1K & 60.7 & - \\
DenseNet \cite{nanni2020insect} & \xmark & ImageNet1K & 61.9 & - \\
GAEnsemble \cite{ayan2020crop} & \xmark & ImageNet1K & 67.1 & - \\
ViT \cite{dosovitskiy2020image} & \xmark & ImageNet1K & 71.6 & 87.7 \\
\hline
MoCo \cite{he2020momentum} & \xmark & 1M-Insect & 70.6 & 88.4 \\
DINO \cite{caron2021emerging} & \xmark & 1M-Insect & 71.5 & 91.4 \\
MAE \cite{he2022masked} & \xmark & 1M-Insect & 72.0 & 91.5 \\
CoCa \cite{yu2022coca} & \cmark & 1M-Insect & 72.8 & 91.1 \\
\hline
\textbf{Insect-Foundation} & \xmark &  1M-Insect & \textbf{73.3} & \textbf{91.6} \\
\textbf{Insect-Foundation} & \cmark &  1M-Insect & \textbf{75.8} & \textbf{92.1} \\
\Xhline{2\arrayrulewidth}
\end{tabular}
}
\vspace{-6mm}
\end{table}

\noindent
\textbf{Effectiveness of Network Backbones}
Table \ref{tab:abl_studies} studies the impact of 
different Vision Transformer backbone sizes, including ViT-small/16, ViT-base/16, and ViT-large/16.
As shown in our results, the powerful backbone carries more improvement. In particular, when changing the Transformer backbone size from small to base, the accuracy score increases by a large margin of $4.3\%$ while the large Transformer backbone improves the accuracy score by $1.1\%$.

\noindent
\textbf{Effectiveness of Attention Pooling}
We evaluate the impact of the attention pooling in the visual representation of the insect images.
As shown in Table \ref{tab:abl_studies}, the Attention Pooling has better representation than the standard classification token computed through transformer layers.
In particular, the top-1 accuracies for the three backbones, i.e., small, base, and large, have been increased from $68.9\%$ to $69.5\%$, from $72.4\%$ to $73.3\%$, and from $73.8\%$ to $74.6\%$.

\noindent
\textbf{Effectiveness of Image-Text Contrastive Loss}
As reported in Table \ref{tab:abl_studies}, the model can understand the insect images better when the model learns to match the images and their descriptions.
In detail, the accuracy scores have been increased by $0.8\%$, $0.9\%$, and $1.3\%$ for the three backbones when applying the Image-Text Contrastive Loss.

\noindent
\textbf{Effectiveness of Description Loss}
The full configuration in Table \ref{tab:abl_studies} shows the experimental results of our model using the Description Loss.
As shown in Table \ref{tab:abl_studies}, the Description Loss helps the model to well-align the information between images and the details of descriptions. Hence, the model can represent the fine-grained features of the insects better.
In particular, the accuracy scores have been improved from $70.7\%$ to $71.5\%$, from $74.2\%$ to $75.8\%$, and from $75.9\%$ to $76.9\%$ for ViT-small/16, ViT-base/16, and ViT-large/16.

\subsection{Comparisons with Prior SOTA Methods}

\begin{table}[!b]
\vspace{-6mm}
\centering
\caption{\textbf{Zero-shot classification results on IP102 Classification benchmark.} The proposed model outperforms prior vision-language pretraining methods.}
\vspace{-3mm}
\resizebox{0.8\linewidth}{!}{
\begin{tabular}{lcc}
\Xhline{2\arrayrulewidth}
Method & Pretrain Data & Accuracy (\%) \\
\hline
CLIP \cite{radford2021learning}  & 1M-Insect     & 41.1    \\
LiT  \cite{zhai2022lit}  & 1M-Insect     & 43.6    \\
CoCa \cite{yu2022coca}  & 1M-Insect     & 45.3    \\
\hline
Insect-Foundation   & 1M-Insect     & \textbf{49.9}    \\
\Xhline{2\arrayrulewidth}
\end{tabular}
}
\label{tab:zero_shot_classification}
\end{table}
\begin{table}[!t]
\centering
\setlength{\tabcolsep}{3pt}
\caption{\textbf{Detection results on IP102 Detection benchmark.} The proposed model outperforms prior pre-training methods.}
\vspace{-3mm}
\resizebox{0.4\textwidth}{!}{
\begin{tabular}{llcccc}
\Xhline{2\arrayrulewidth}
\textbf{Method} & 
\textbf{Backbone} & 
\makecell{\textbf{Pre-train} \\ \textbf{Data}}   & 
\makecell{\textbf{AP} \\ \textbf{(\%)}}   & 
\makecell{$\textbf{AP}^{\textbf{.50}}$ \\ \textbf{(\%)}} & 
\makecell{$\textbf{AP}^{\textbf{.75}}$ \\ \textbf{(\%)}} \\
\hline
FRCNN \cite{ren2015faster}  & VGG-16 \cite{simonyan2014very} & ImageNet1K & 21.1 & 47.9 & 15.2 \\
FPN \cite{lin2017feature}   & ResNet-50 \cite{he2016deep} & ImageNet1K & 28.1  & 54.9 & 23.3  \\
SSD300 \cite{liu2016ssd}    & VGG-16 \cite{simonyan2014very}  & ImageNet1K & 21.5 & 47.2 & 16.6 \\
RefineDet \cite{zhang2018single} & VGG-16 \cite{simonyan2014very} & ImageNet1K  & 22.8 & 49.0 & 16.8 \\
YOLOv3 \cite{redmon2018yolov3}    & DarkNet-53 \cite{redmon2018yolov3}  & ImageNet1K & 25.7 & 50.6 & 21.8 \\
FPN \cite{lin2017feature} & ViT \cite{dosovitskiy2020image} & ImageNet1K & 32.8 & 54.7 & 35.0 \\
\hline
FPN \cite{lin2017feature} & MoCo \cite{he2020momentum} & 1M-Insect & 33.6 & 56.1 & 35.3 \\
FPN \cite{lin2017feature} & DINO \cite{caron2021emerging} & 1M-Insect & 34.0 & 55.8 & 37.1 \\
FPN \cite{lin2017feature} & MAE \cite{he2022masked} & 1M-Insect & 34.7 & 58.4 & 37.8 \\
\hline
FPN \cite{lin2017feature} & Insect-Foundation & 1M-Insect & \textbf{36.6} & \textbf{59.1} & \textbf{40.3} \\
\Xhline{2\arrayrulewidth}
\end{tabular}
}
\label{tab:ip102_detection}
\vspace{-7mm}
\end{table}

\noindent
\textbf{Insect Classification Tasks.}
We fine-tune the linear layer with our pre-trained model on the IP102 dataset \cite{wu2019ip102} 
for the classification task.
As shown in Table \ref{tab:ip102_classification}, our model outperforms deep learning models \cite{krizhevsky2012imagenet,szegedy2015going,simonyan2014very,he2016deep,dosovitskiy2020image} pre-trained on ImageNet \cite{deng2009imagenet} by a large margin.
Compared to other pre-training methods \cite{he2020momentum,caron2021emerging,he2022masked,yu2022coca} on the proposed 1M-Insect dataset, our model shows better performance for both training without and with insect descriptions of $73.3\%$ and $75.8\%$, respectively.
It is shown that the proposed approach has a better visual representation of insect images than the prior pre-training methods on the same dataset.

\begin{figure}[!t]
\begin{center}
\includegraphics[width=0.85\linewidth]{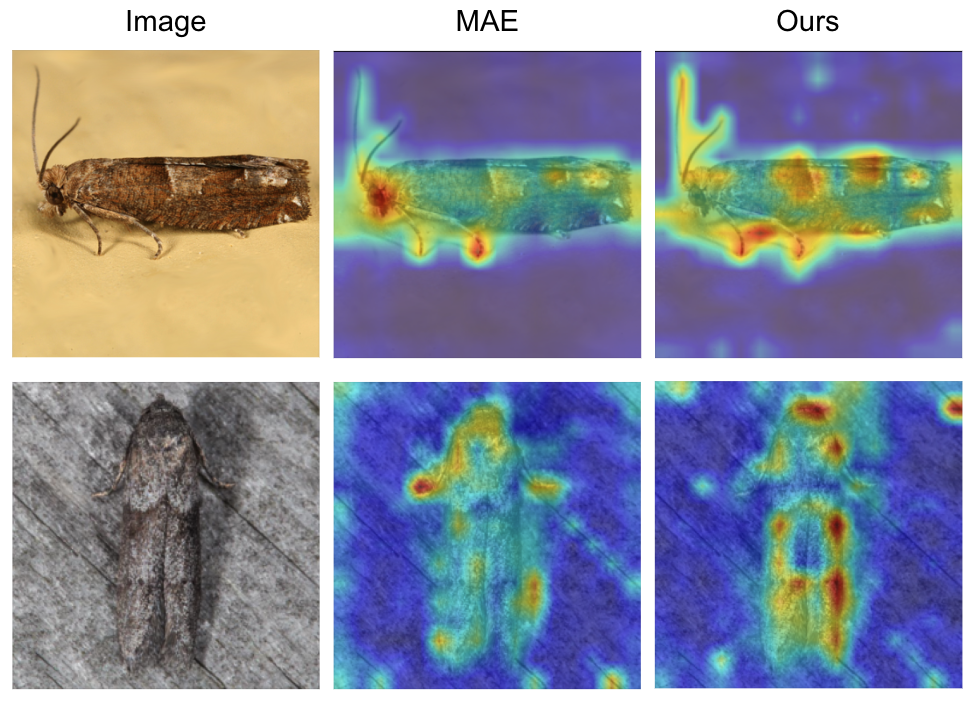}
\end{center}
\vspace{-7mm}
\caption{\textbf{Attention Visualization.} Compared to MAE \cite{he2022masked}, our model is robust to small details of insect images. The model can focus on the small textures of the insect, even if the texture is the same as the background (bottom images). \textbf{Best viewed in color.}}
\label{fig:attention_visualization}
\vspace{-7mm}
\end{figure}

\noindent
\textbf{Visualization Results} Fig. \ref{fig:attention_visualization} visualizes the attention maps of our model compared to MAE \cite{he2022masked} pre-trained on the proposed dataset.
Since the textures are similar to the background, it is hard for MAE to focus on the small details of the insect.
On the contrary, our model can detect the key features, i.e., the textures and the limbs, of the insects.

\noindent
\textbf{Zero-shot Insect Classification.}
We evaluate the performance of our model on the IP102 dataset \cite{wu2019ip102} in a zero-shot manner.
In detail, a description corresponds to each species to make the text encoder extract more semantic information about each species.
Then, for each insect image, we use the image encoder to extract global features and compare them to each description feature to predict the insect species.
Table \ref{tab:zero_shot_classification} reports the results of zero-shot classification on the IP102 Classification benchmark.
Our model outperforms prior image-text pre-training methods \cite{radford2021learning,zhai2022lit,yu2022coca} at an accuracy of $49.9\%$.
It shows that our model has well-alignment between the insect image and its description.

\noindent
\textbf{Insect Detection Tasks.}
As shown in Table \ref{tab:ip102_detection}, we train a Faster R-CNN model \cite{ren2015faster} on the IP102 Detection dataset with the ViT backbone adapted for FPN \cite{lin2017feature}.
Compared to models pre-trained on ImageNet \cite{deng2009imagenet}, our model achieves SOTA results with an average precision of $36.6\%$ and $\text{AP}^{.50}$ of $59.1\%$ higher than the same backbone pre-trained on ImageNet \cite{deng2009imagenet} having AP of $32.8\%$ and  $\text{AP}^{.50}$ of $54.7\%$.
Compared to other self-supervised methods \cite{he2020momentum,caron2021emerging,he2022masked}, our model achieves higher precision.
Thus, our model focuses on the features of insects better than prior methods.

\section{Conclusions}

This paper has introduced a new large-scale Insect-1M dataset that supports the development of the Insect Foundation Model in precision agriculture. Our proposed dataset includes a large diversity of insect species and multi-level labels of taxonomy. In addition, Insect-1M consists of detailed descriptions of insects that support vision-language insect model training. 
Then, to improve the micro-feature modeling of our insect foundation model, we introduce a new Patch-wise Relevant Attention mechanism and Description Consistency loss to learn the details of insects. 
Our experimental results have illustrated the effectiveness and significance of our Insect-1M and Insect Foundation Model.

\noindent
\textbf{Limitations} This study used a specific network design and learning hyper-parameter to support our hypothesis. However, our approach potentially consists of several limitations related to the design of our Patch-wise Relevant Attention mechanism, where the patches of background and foreground are equally treated. It could result in difficulty in learning the different features of insects.
This limitation will further motivate future research to improve the Insect Foundation Model and Micro-feature Modeling.

\noindent  \textbf{Acknowledgment.} This work is partly supported by NSF DART, NSF SBIR Phase 2, and JBHunt Company. We also acknowledge the Arkansas High-Performance Computing Center for GPU servers and Jesse Ford for dataset tasks.

{
    \small
    \bibliographystyle{ieeenat_fullname}
    \bibliography{main}
}

\end{document}